\documentclass[10pt,twocolumn,letterpaper]{article}

\usepackage{iccv}
\usepackage{times}
\usepackage{epsfig}
\usepackage{graphicx}
\usepackage{amsmath}
\usepackage{amssymb}
\usepackage{booktabs}

\usepackage[breaklinks=true,bookmarks=false]{hyperref}

\iccvfinalcopy 

\def\iccvPaperID{****} 

\ificcvfinal\fi

\begin{document}

\title{AU-Supervised Convolutional Vision Transformers for Synthetic Facial Expression Recognition}

\author{Shuyi Mao, Xinpeng Li, Junyao Chen, Xiaojiang Peng\thanks{Correspondence: pengxiaojiang@sztu.edu.cn}\\
Shenzhen Technology University, Shenzhen, China\\
}


\maketitle

\ificcvfinal\thispagestyle{empty}\fi

\begin{abstract}
The paper describes our proposed methodology for the six basic expression classification track of Affective Behavior Analysis in-the-wild (ABAW) Competition 2022. In Learing from Synthetic Data(LSD) task, facial expression recognition (FER) methods aim to learn the representation of expression from the artificially generated data and generalise to real data. Because of the ambiguous of the synthetic data and the objectivity of the facial Action Unit (AU), we resort to the AU information for performance boosting, and make contributions as follows. 
First, to adapt the model to synthetic scenarios, we use the knowledge from pre-trained large-scale face recognition data. Second, we propose a conceptually-new framework, termed as AU-Supervised Convolutional Vision Transformers (AU-CVT), which clearly improves the performance of FER by jointly training auxiliary datasets with AU or pseudo AU labels. Our AU-CVT achieved F1 score as $0.6863$, accuracy as $0.7433$ on the validation set.
The source code of our work is publicly available online: \href{https://github.com/msy1412/ABAW4}{https://github.com/msy1412/ABAW4}
\end{abstract}

\section{Introduction}
Facial Expressions Recognition (FER) aims to identify people's emotion in a facial image. It plays a great role in various applications including surveillance \cite{clavel2008fear}, games \cite{pioggia2005android}, educations \cite{yang2018emotion}, medical treatments \cite{pioggia2005android} and marketing \cite{ren2012linguistic}. In the past decade, more and more scholars in computer vision have 
devoted to this task and contributed a lot \cite{corneanu2016survey, noroozi2018survey, rouast2019deep, zhang2018facial, li2020deep}. However, analyzing affect behaviors in the wild is a challenging affect behaviors in the wild is a challenge. In Aff-Wild\cite{zafeiriou2017aff} and AffWild2\cite{kollias2018aff,kollias2019expression,kollias2022abaw}, authors proposed the very first video-based affect behavior database, in which it contains 7 basic expressions, 2D valence-arousal, and 12 facial action units(AUs) annotations, to facilitate the society to solve these problems. Several multi-task approaches have been introduced to recognize these affective behavior\cite{kollias2019face,kollias2021affect,kollias2021distribution}. For the Learning from Synthetic Data Challenge in ABAW Competition 2022, the goal is to create a system that learns to recognise the six basic expressions (anger, disgust, fear, happiness, sadness, surprise) from artificially generated data (i.e., synthetic data) and generalise its knowledge to real-world (i.e., real) data\cite{kollias2022abaw}. Theses synthetic data are some specific cropped images of Aff-Wild2 database, which have been used in a facial expression manipulation manner\cite{kollias2020deep,kollias2020va,kollias2018photorealistic}.

Psychological studies show that Facial Action Unit (AU) is an objective and common standard to describe physical expression of emotions \cite{ekman1997face}. For instance, $AU12$ and $AU6$ indicate lip corner puller and cheek raiser, which relate to happiness; $AU1$, $AU4$, and $AU6$ denote inner brow raiser, brow lower, and cheek raiser, which relate to sadness. On the one hand, the faces with same emotion show different AU; on the other hand, the AU is objective description of facial motion. It suggests that emotions from different databases (e.g.,synthetic data and real data) share unbiased and universal appearance through AU. In other word, AU provides possibility to guide FER model to learn the crucial changing of face, which help to generalise the features it learns from synthetic data to real-world scenario. Therefore, Inspired by we take advantage of AU to make auxiliary datasets useful.

We propose a conceptually-new network architecture, termed as AU-Supervised Convolutional Vision Transformer (AU-CVT). The AU-CVT mainly takes AUs or pseudo AUs as a bridge between target and auxiliary datasets since the image-to-AU procedure is objective.

Specifically, the AU-CVT consists of three vital branches, namely an Emotion branch (Exp-branch), and an AU branch. It first takes as input the images from both target and auxiliary datasets, and then encodes them into middle-level feature maps by the B-branch, and finally feeds them into either the Emotion branch for facial expression classification or the AU branch for AU recognition. For the Emotion branch, a convolution based ViT (CNN-ViT) applies Transformers on feature maps. For the AU branch, we elaborately design several patch-splitting strategies for feature aggregation, motivated by the fact that AUs are mainly defined by local information. The Emotion branch handles the target dataset, and the AU branch only deals with the auxiliary datasets which own AU or pseudo AU labels, and is expected to boost the feature learning of the Emotion branch leading to better expression recognition. 

With a carefully pre-trained backbone, our AU-CVT further improves the baseline by around 0.18 and achieves $F1$ as 0.6863, accuracy as 0.7433.

\section{Related work}
Numerous scholars in computer vision have devoted to FER and contributed a lot of works, including algorithms~\cite{kollias2017recognition,wang2020guided,wang2020region,wang2021face}. 
This paper propose AU-Supervised Convolutioanal Vision Transformer to address how to boost FER performance by leveraging cross-domain datasets. Therefore, we will show FER related works of vision transformers and AU assistance. 

\textbf{Vision Transformer}.
Vision Transformers have been proved effective in image classification, detection, and segmentation \cite{dosovitskiy2020image, 2021CrossViT, wang2021pyramid, heo2021rethinking, liu2021swin}. Recently, some works began to introduce it to the FER task. \cite{ma2021facial} combines two kinds of feature maps generated by two-branch CNNs, then feeds the fused feature into Transformers to explore relationships between visual tokens. \cite{xue2021transfer} leverages Vision Transformers upon CNNs to learn rich relations among diverse local patches between two attention droppings.

\textbf{AU Assistance}.
Recent works began to utilize AU information to boost FER since AUs are related to emotions \cite{ekman1978facial}. For example, \cite{deng2020multitask} presents a distillation strategy to learn from incomplete labels to boost the performance of facial action unit detection, expression classification, and valence-arousal estimation; \cite{pu2021expression} exploits the relationships between AU and expression and chooses useful AU representations for FER; \cite{chen2021understanding} leverages AUs to understand and mitigate annotation bias.

\section{Methodology}
To recognize crucial changing in human faces, we propose an AU-Supervised Convolutional Vision Transformer (AU-CVT) for joint dataset training in facial expression recognition.

\begin{figure*}
\begin{center}
\includegraphics[width=0.85\textwidth]{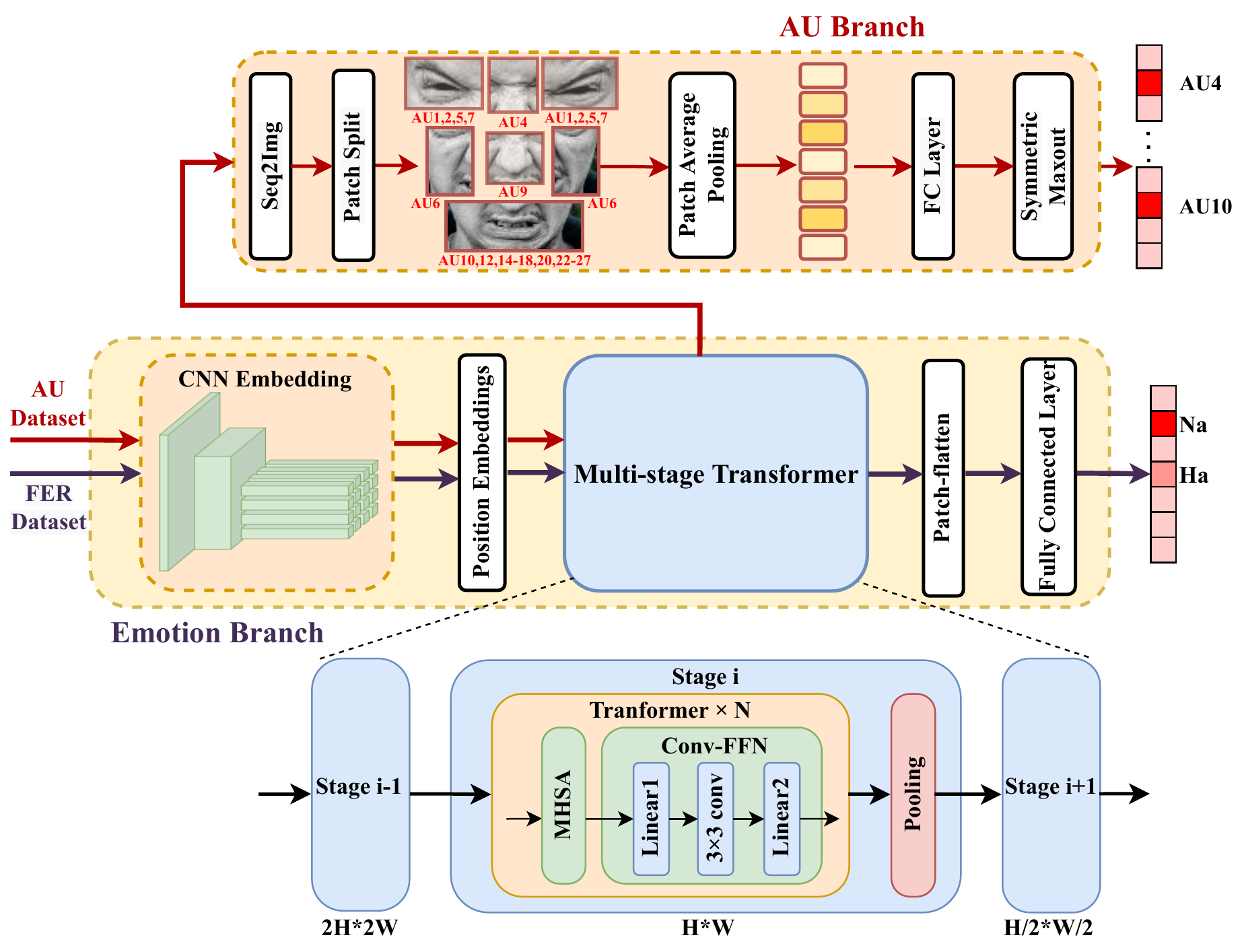}
\end{center}
   \caption{Pipeline of the proposed model.}
\label{fig:pipeline}
\end{figure*}

\subsection{AU-CVT Pipeline}
As shown in Figure \ref{fig:pipeline}, the AU-CVT consists of a ViT emotion branch and an extended AU branch. The ViT emotion branch takes charge of the IM-to-emotion procedure, adopting occlusion-robust Vision Transformers. The extended AU branch is responsible for the IM-to-AU procedure, using shared linear projections and Transformers.

We train a target dataset and an auxiliary dataset simultaneously. The target dataset is processed by the ViT emotion branch for the emotion classification. The auxiliary dataset is extracted by the extended AU branch for the AU classification, which is objective between the target dataset and the auxiliary dataset. In this way, we get more training samples due to joint training while avoiding dataset bias due to an AU classification. 

\subsection{Emotion Branch}
As shown in Figure \ref{fig:pipeline}'s middle part, the CVT emotion branch mainly consists of the CNN Embedding module and the Multi-stage Transformer. We employ the first three stages of ResNet50 as backbone to generate embeddings with the size of $\frac{H}{R}\times\frac{W}{R}\times C$ in CNN Embedding module, where R is the down-sampling rate of the backbone and C is the channel of the output of the backbone. Then the multi-stage Transformer is employed on the feature maps generated by IR50. As the layers of Transformers deepen, we gradually increase the channel size while decreasing the spatial size in Transformers, like previous works\cite{wang2021pyramid}\cite{heo2021rethinking}\cite{liu2021swin}. We name this architecture . Before inputting the first Transformer block, the feature maps are split into non-overlapping patches, like simple ViT. In implement, we set patch size as $1\times 1$ and the dimension of each patch embedding is 256. Unlike the above simple version, the class token embedding is discarded. Moreover, the number of tokens in the Multi-stage Transformer is reduced by patch merging layers as the network gets deeper. In the end of each stage, the pooling layer would concatenate the tokens of each group of $2\times2$ neighbour patches as one token by increasing its dimension from $C$ to $2C$. So the size of feature maps in the end of the $K^{th}$ stage of Transformer is $\frac{H_{f}}{2K}\times \frac{W_{f}}{2K}\times 2KC_{f}$, where $H_{f}$ and $W_{f}$ are the height and width of feature maps from ResNet50.

In order to model the complex interactions among all elements of the patch embeddings, we input $Z^{0}$ to the Transformer encoder. The Transformer encoder calculates the weights of embeddings $Z^{0}$ through multi-head self-attention (MHSA).

To employ connectivity of local regions into ViT, we add a depth-wise convolution in feed-forward networks (Conv-FFN) after MHSA module.
In the Conv-FFN module, we rearrange the sequence of tokens into a 2D lattice and convert it to 2D feature maps. The number of channels of the feature maps increases firstly, and then a depth-wise convolution with a kernel size of $3\times3$ performs on them. Finally, we restore the channels of feature maps and flatten them into sequences to the initial dimension.The local convolution enhances the representation correlation with neighbouring 8 tokens. 

It is worth noting that the class token is not used in the Multi-stage Transformer. So we flatten all the patch embeddings in the Patch-flatten layer then implement a fully connected layer instead of an average pooling to output final classification.
Flatten operation and FC-layers can retain all spacial information of feature maps of each channel and the strong relationships among the features of different areas, which is more vital to facial expression recognition than any other computer vision task.

\subsection{AU Branch}
The AU branch processes the auxiliary dataset for AU recognition, as is shown in Figure \ref{fig:pipeline}’s upper part. The AU branch and the emotion branch share the former operation and transformer blocks. Then, a $Seq2Img$ operation transforms the $C\times N$ patch tokens to $C\times {H_f}\times {W_f}$ 2D feature maps. After that, we split the 2D feature maps into several overlapped patches according to the defined location of AUs.

The upper part of Figure \ref{fig:pipeline} plots our AU-patch splitting strategy. We split the feature maps into 7 patches: the left eye, right eye, left cheek, right cheek, between-eyebrow, nose and mouth. The left and right-eye patches are responsible for AU1, AU2, AU5, and AU7. Likewise, the left and right-cheek patches are both responsible for AU6. The between-eye and nose patches aim to recognize AU4 and AU9, respectively. In particular, the mouth patch is the largest part among these patches, nearly half of the aligned face. Because 14 kinds of AU(10, 12, 14, 15, 16, 17, 18, 20, 22, 23, 24, 25, 26, 27) are related to the mouth patch. All the patches are overlapped because of the necessity of redundant features. 

Then, we calculate the patch average pooling of each patch to obtain seven $C\times1$ vectors. These vectors are fed into the fully-connected layers to output the recognition results of AU. Since the left-eye patch and right-eye patch are symmetric and responsible for the same AUs, we select the maximum of their output in the symmetric maxout layer, and so do the pair of the left-cheek patch and the right-cheek patch. The formula of symmetric maxout is as follows:
$$
Y_{AU_{i}}=max(Y_{AU_{left}},Y_{AU_{right}}),
$$
where, $AU_{left}$ and $AU_{right}$ mean AUs in left and right faces.

After the symmetric maxout layer, all the predictions are used for calculating the loss with AU labels.

\subsection{Loss Function}
Let $z_{FER}$ be the logits of the final output of the emotion Branch, $z_{AU}$ the logits of the AU Branch. The former inputs the SoftMax function($\psi$) while the latter inputs the Sigmoid function($\phi$). We denote the two coefficient values $\alpha$ and $\beta$ to balance the cross-entropy loss ($L_{CE}$) of FER and binary cross-entropy loss ($L_{BCE}$) of AU detection. $y_{FER}$ and $y_{AU}$ are the ground truth labels from FER dataset and AU Detection dataset respectively. The overall loss function of AU-CVT is as follow:
$$
\mathcal{L_{\text {global}}}=\alpha*L_{\text {CE}}(\psi(z_{FER}),y_{FER}) +\beta*L_{\text {BCE}}(\phi(z_{AU}),y_{AU})
$$
\section{Experiment and Result}
\subsection{Dataset}
To evaluate our method, we use the LSD dataset as target dataset and RAF-AU~\cite{Yan_2020_ACCV} as the auxiliary dataset.

\textbf{LSD}~\cite{kollias2022abaw}Learning from Synthetic Data Challenge in the ABAW4 consists of 277,251 images that contain annotations in terms of the 6 basic expressions (anger, disgust, fear, happiness, sadness, surprise). The validation and test sets of this Challenge are real images of the Aff-Wild2. The synthetic images are for training, which have been generated from subjects of the validation set, but not of the test set. Both the validation sets and test sets only contain real data of the Aff-Wild2. However the annotation of the validation sets are available to the participants. The test set without annotations is given to the participants. We mainly report the $F1$ score and the accuracy. 

\textbf{RAF-AU}~\cite{Yan_2020_ACCV}. The RAF-AU is an extended dataset of RAF-ML collected from the Internet with blended emotions. It varies in subjects' identity, head poses, lighting conditions and occlusions.
The face images in RAF-AU are FACS-coded by two experienced coders independently. It contains 4601 real-world images with 26 kinds of AUs being annotated,including AU1, 2, 4, 5, 6, 7, 9, 10, 12, 14, 15, 16, 17, 18, 20, 22, 23, 24, 25, 26, 27, 28, 29, 35, 43. In our experients, we only use the first 21 AUs, which have a strong relationship with expressions. We use RAF-AU as a auxiliary dataset.

\subsection{Implementation Detail}
For our AU-CVT, ResNet50 is used as our CNN Embedding and initialized by the pre-trained weights on the MS-Celev-1M\cite{guo2016ms}. Since the LSD dataset doesn’t contain AU annotations, we utilize an extra AU datasets, the RAF-AU, or generate pseudo AU labels for the LSD dataset by the AU detector OpenFace\cite{baltrusaitis2018openface}.
We feed the FER labels and the AU labels into our network at the same time to train the emotion Branch and the AU Branch. For each image, we resize the image with a short edge of 112 for training. For data augment, we use Random-Horizontal Flip, Random-Grayscale and Gaussian Blur for all datasets. The optimizer is SGD, the weight decay is 5e-4, and the batch size is 512 for the LSD dataset. In the training phase, the learning rate of our methods is initialized as 0.0001, and we use a linear learning rate warm-up of 3 epochs and cosine learning rate decay of 7 epochs. All the experiments are implemented by Pytorch with 4 NVIDIA V100 GPUs. The default $\alpha$ and $\beta$ are 1.0 and 1.0, respectively. 




\begin{table}[tp]
\center
\caption{\textcolor[rgb]{0.00,0.00,0.00}{Comparison on the LSD dataset. CVT denotes the ViT with CNN Embedding.}}
\begin{tabular}{cccc}
\toprule
Method       & Auxiliary Data     & F1      & Acc   \\\midrule
ResNet50(baseline)  &/  &50       &/ \\
ViT         &/      &59.83       &67.02  \\
CVT     &/      &65.02          &73.98\\
AU-CVT      &RAF-AU    &67.10      &73.25\\
AU-CVT      &LSD-OpenFace   &68.63  &74.33\\
\bottomrule
\end{tabular}
\label{tab:compare}
\end{table}

\subsection{Result and Discussion}
As Table \ref{tab:compare} showed, ResNet50 pretrained on ImageNet sets the baseline with F1 of 50\%. ViT pretrained on ImageNet improves the baseline 9.38\%. CVT which denotes the ViT with CNN Embedding, outperforms the ViT by 5.19\%.

To supervise the CVT to aware the representation of AUs, we use the AU Branch to process AU labels from AU datasets or pseudo AU labels of LSD generated by OpenFace\cite{baltrusaitis2018openface}. Although RAF-AU dataset has 26 kinds of AU labels, not all AUs contribute to facial expression. Therefore, we select the AU labels which are most relative to facial expression. 
For RAF-AU, we just keep 21 kinds of AU, including AU1, 2, 4, 5, 6, 7, 9, 10, 12, 14, 15, 16, 17, 18, 20, 22, 23, 24, 25, 26, 27. In the training phase, we feed the images from both the LSD and the RAF-AU into AU-CVT, and calculate the final loss with the emotion labels of LSD and AU labels of RAF-AU. The AU Branch guides the Emotion branch to learn a good representation of AU to improve the CVT by 2.08\%. 
Due to the lack of AU labels in LSD dataset, we use OpenFace\cite{baltrusaitis2018openface} to generate 16 kinds of AU annotations for the LSD, including AU1, 2, 4, 5, 6, 7, 9, 10, 12, 14, 15, 17, 20, 23, 25, 26.
During the training, we feed the images of the LSD into the AU-CVT and calculate the expression loss and the pseudo AU loss. AU-CVT achieve the best performance of 68.63\%.


\section{Conclusion}
In this work, we propose the AU-CVT for synthetic facial expression recognition. The Emotion Branch enables the model to extract different scale features and enhance the ability of learning global and local information. AU branch utilizes extra AU annotations to sense precisely the crucial changes on face. 
Based on our experiment results, we have observed that the AU-CVT achieved $F_1$ as 0.6863 and accuracy as 0.7433 on the validation set.

{\small
\bibliographystyle{plain}
\bibliography{Reference_BibTex}
}

\end{document}